\documentclass[runningheads]{llncs}
\usepackage[T1]{fontenc}
\usepackage{graphicx}
\usepackage{color}
\definecolor{myred}{rgb}{.8,.0,.0}

\newcommand{\chex}{CheXpert}
\newcommand{\cxr}{NIH-CXR14}
\newcommand{\padchest}{PadChest}
\def\eg{\textit{e.g.}~}

\usepackage{booktabs}
\usepackage{caption}
\usepackage{subcaption}
\usepackage{framed,multirow}
\usepackage{amsmath}

\newcommand{\JimSan}{Jim\'{e}nez-S\'{a}nchez}

\usepackage[breaklinks=true,bookmarks=false]{hyperref}

\usepackage{array}
\usepackage{tabulary}
\usepackage{bm}
\usepackage{orcidlink}

\newcolumntype{K}[1]{>{\centering\arraybackslash}p{#1}}
\usepackage{tabstackengine}
\TABstackMath

\begin{document}
\title{Augmenting Chest X-ray Datasets with Non-Expert Annotations}
%
%
\author{Veronika Cheplygina \inst{1}\orcidlink{0000-0003-0176-9324} \and
Cathrine Damgaard \inst{1} \and
Trine Naja Eriksen \inst{1} \and \\
Dovile Juodelyte \inst{1}\orcidlink{0000-0002-6195-1120} \and
Amelia \JimSan \inst{1}\orcidlink{0000-0001-7870-0603}\thanks{Corresponding author: Amelia \JimSan, amji@itu.dk.}
}
\authorrunning{Cheplygina et al.}
%
\institute{IT University of Copenhagen, Copenhagen, Denmark \\
\email{\{vech,amji\}@itu.dk}}

\maketitle              

\begin{abstract}

The advancement of machine learning algorithms in medical image analysis requires the expansion of training datasets. A popular and cost-effective approach is automated annotation extraction from free-text medical reports, primarily due to the high costs associated with expert clinicians annotating medical images, such as chest X-rays. However, it has been shown that the resulting datasets are susceptible to biases and shortcuts. Another strategy to increase the size of a dataset is crowdsourcing, a widely adopted practice in general computer vision with some success in medical image analysis.  In a similar vein to crowdsourcing, we enhance two publicly available chest X-ray datasets by incorporating non-expert annotations. However, instead of using diagnostic labels, we annotate shortcuts in the form of tubes. We collect 3.5k chest drain annotations for \cxr{}, and 1k annotations for four different tube types in \padchest{}, and create the Non-Expert Annotations of Tubes in X-rays (NEATX) dataset. We train a chest drain detector with the non-expert annotations that generalizes well to expert labels. Moreover, we compare our annotations to those provided by experts and show ``moderate'' to ``almost perfect'' agreement. Finally, we present a pathology agreement study to raise awareness about the quality of ground truth annotations. We make our dataset available on Zenodo at \url{https://zenodo.org/records/14944064} and our code available at \url{https://github.com/purrlab/chestxr-label-reliability}.

\end{abstract}

\section{Introduction} \label{sec:intro}
The use of machine learning (ML) algorithms is nowadays standard for medical image analysis, yet there are still issues with the size and representativeness of the datasets used to train the algorithms. While it is clear that we need large and high-quality datasets for the \emph{validation} of the algorithms, there have been various efforts to reduce the number of expert labels needed during \emph{training}. These efforts include methods development (\eg{}semi-supervised \cite{cheplygina2019not} or active learning \cite{Budd2021ActiveLearning}) or alternative labeling strategies, such as label extraction via natural language processing or crowdsourcing, we focus on the latter in this paper. 

One popular strategy consists of creating larger datasets based on the automatic extraction of labels. For example, CheXpert~\cite{irvin2019chexpert} and \cxr{}~\cite{chest-xray8} extracted pathology labels from the free text clinical reports, and \padchest{}~\cite{BUSTOS2020101797} combined expert diagnostic labels with ML-extracted annotations. However, CheXpert and \cxr{} datasets have been shown to suffer from bias \cite{gichoya2022ai,larrazabal2020gender,seyyed2020chexclusion} or shortcuts \cite{JimenezSanchez2023Shortcuts,oakden2020hidden}. 

In general computer vision tasks, crowdsourcing of non-expert labels has been a popular solution \cite{kovashka2016crowdsourcing}, with some successes reported in medical imaging as well \cite{orting2020survey}. For chest X-ray, Filice et al.~\cite{filice2020crowdsourcing} collected crowdsourced annotations from six board-certified radiologists for the ``pneumothorax'' class in \cxr{}. In chest CT scans, non-experts have assessed the similarity of lung tissue \cite{orting2017crowdsourced}, and also annotated pathological patterns, \cite{oneil2017crowdsourcing}, nodules \cite{boorboor2018crowdsourcing} and airways \cite{cheplygina2021crowdsourcing}. 

While the previous work on chest X-rays focused on relabeling pathologies, we extend annotation types by labeling ``shortcuts'' in two datasets. We refer to shortcuts as spurious correlations between artifacts in images and diagnostic labels \cite{banerjee2023shortcuts,jimenez2025picture}. Our method is close in spirit to crowdsourcing, but rather than extending the dataset size in terms of diagnostic labels, we annotate possible shortcuts, increasing the richness of the annotations.
Our contributions are:  
\begin{itemize}
    \item We introduce the Non-Expert Annotations of Tubes in X-rays (NEATX) dataset\footnote{\scriptsize{\url{https://zenodo.org/records/14944064}}}, which enriches two publicly available chest X-ray datasets with annotations from non-experts. We collect 3.5k drain annotations for \cxr, and 1k annotations for four different tube types in \padchest.
    \item We present a pathology agreement study on different labels created for \cxr{} to raise awareness about potential limitations of the available ground truth annotations.
    \item We train a chest drain detector with our non-expert annotations and validate the performance showing that it generalizes well to expert drain annotations for \cxr{}.
    \item We train a tube classifier on \padchest{} and show that the non-expert annotations achieve similar, or better, performance than the extracted by ML.
    \item We make our code for the experiments available\footnote{\scriptsize{\url{https://github.com/purrlab/chestxr-label-reliability}}}. 
\end{itemize}

\begin{figure}[t]
    \centering
    \includegraphics[width=0.9\columnwidth]{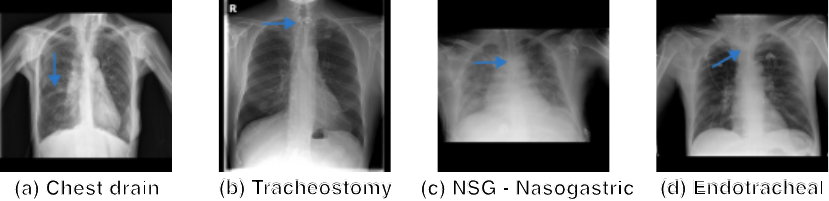}
  \caption{Tube examples in \padchest{} dataset.}
  \label{fig:chest-xray-tubes}
\end{figure}

\section{Related work} \label{sec:related}
\subsection{Experts-provided annotations}
The cost of annotating medical image datasets influences algorithm development. In \cite{kim2022did}, this cost is defined as a function of three factors: \emph{quantity}, \emph{quality}, and \emph{granularity} of the annotations. They found that cost-efficient annotations provide great value for training multi-label classification and segmentation in frontal chest X-ray, and that combining these annotations with a limited number of high-cost labels lead to competitive models at a much lower expense.

Building on a similar concept, \cite{filice2020crowdsourcing,radsch2021radiologist} proposed techniques to automatically identify chest X-ray samples for expert validation. Mislabeled instances in a dataset might not only impact the performance of ML models negatively but also endanger the explainability and reliability of the predictions. Filice et al.~\cite{filice2020crowdsourcing} explored ML methods for generating annotations in chest X-rays for expert review, with multiple experts labeling pneumothorax in the \cxr{} dataset. R\"{a}dsch~et al.~\cite{radsch2021radiologist} proposed a ML approach for the automatic detection of mislabeled instances in \chex{} dataset. Their method identified 7.4\% of mislabeled Cardiomegaly cases. Moreover, they validated mislabeled cases through a blind study with a professional radiologist, supporting ongoing data quality efforts. In contrast, our approach focuses on enhancing the dataset's richness with a different type of annotation (shortcuts), and does not rely on clinical expertise.

\subsection{Annotations from non-experts}
Crowdsourcing is popular in computer vision \cite{kovashka2016crowdsourcing}, but medical imaging presents several challenges that make the transferability of that success more difficult. Some approaches have successfully leveraged non-expert knowledge to gather annotations \cite{albarqouni2016aggnet,keshavan2018combining,sharma2017crowdsourcing}. \O{}rting et al. \cite{orting2017crowdsourced} survey 57 papers applying crowdsourcing to medical imaging. They conclude that most published studies find that crowdsourcing is a viable solution, often for segmentation tasks which might be more intuitive to non-experts. Focusing on lung images, non-experts have annotated pathological patterns, \cite{oneil2017crowdsourcing}, nodules \cite{boorboor2018crowdsourcing} and airways \cite{cheplygina2021crowdsourcing}, or assessed similarity of lung tissue \cite{orting2017crowdsourced}, all in chest CT scans, reporting added value to ML algorithms. 

One of the main limitations of crowdsourcing is that is non-trivial how to setup a crowdsourcing project for medical images, and the surveyed papers often do not provide enough details on how this is done. For example, papers often do not report how the annotators were recruited, incentivized, and trained to perform the annotation tasks.

Our work is similar in spirit to crowdsourcing, however, instead of enlarging the dataset with additional diagnostic labels, we annotate potential shortcuts, enhancing the depth of the annotation information. This alternative form of annotation allows us to assess whether shortcut issues may be present in medical image diagnosis tasks in chest X-rays.

\section{Methods} \label{sec:methods}
\begin{figure}[t]
    \centering
    \includegraphics[width=0.8\columnwidth]{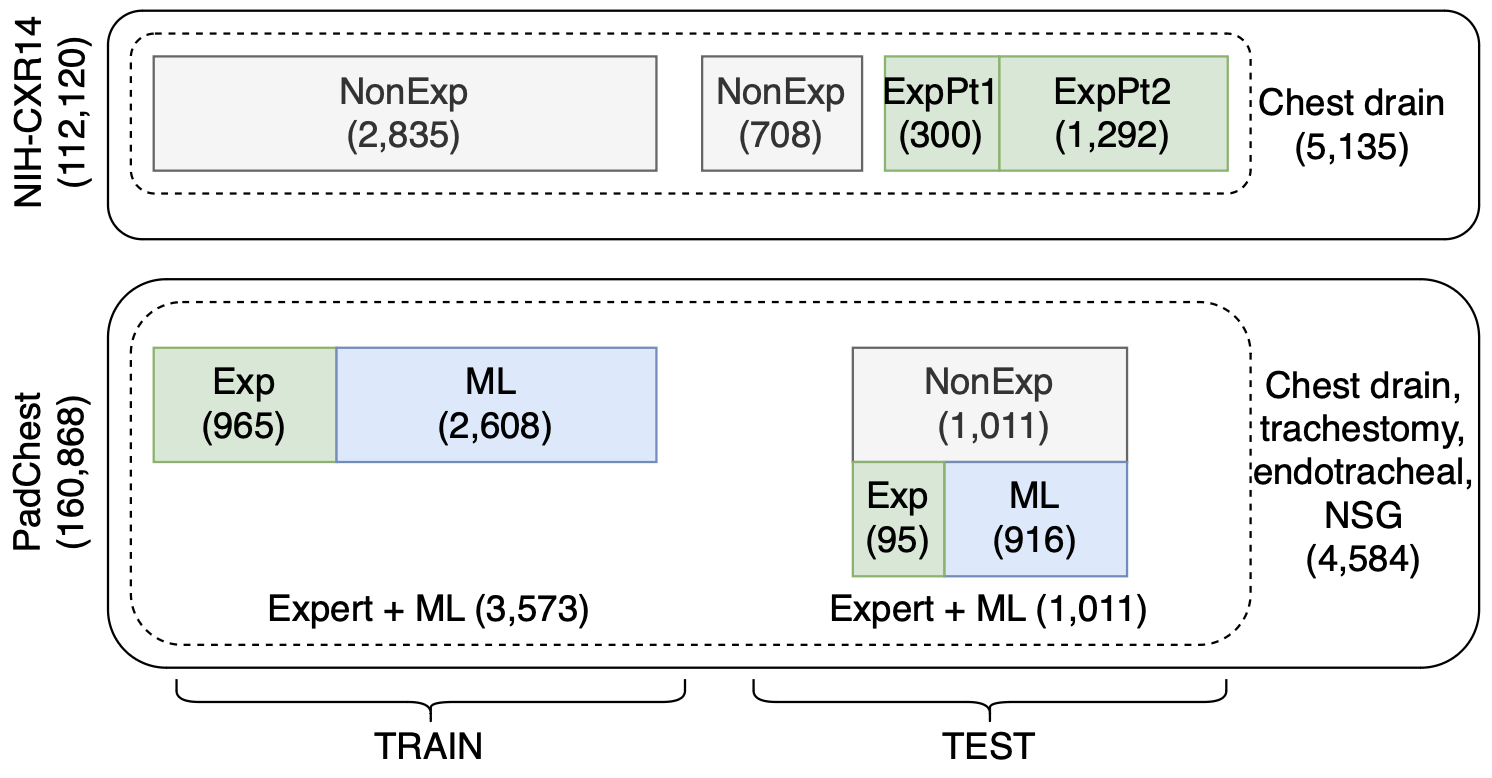}
    \caption{Train and test subsets for \cxr{} (top) and \padchest{} (bottom). \texttt{ExptPt1, ExpPt2} refer to the annotations provided by an external expert at the beginning and end of our study, respectively. \texttt{Exp} are the original \padchest{} labels from clinicians looking at the reports.}
    \label{fig:subsets}
\end{figure}

\subsection{Datasets} \label{subsec:data}
We create the \textbf{Non-Expert Annotations of Tubes in X-rays (NEATX)} dataset enriching the annotations of two publicly available chest X-ray datasets: \cxr{}~\cite{chest-xray8} and \padchest{}~\cite{BUSTOS2020101797}. We release NEATX on Zenodo, a platform offering persistent storage and a Digital Object Identifier (DOI). 

\textbf{\cxr{}} originally has 112,120 images from 30,805 patients associated with 14 classes corresponding to different pathologies. We extracted 5,302 images labeled as pneumothorax cases. The annotations provided were text-mined from reports and may not always reflect the visual contents of the images \cite{oakden2020exploring}.

For \cxr{}, multiple additional label sets are available. We investigate the following four: BBox dataset \cite{chest-xray8}, released alongside \cxr; two GCS datasets---GCS16L \cite{CHX14_alllabels} and GCS4L \cite{CHX14_fourlabel}---collected from two independent studies; and the RSNA set \cite{RSNA}, designed to evaluate performance on pneumonia detection. Additional details about these datasets are provided in Table~\ref{tab:datasets-labels}.

In addition, we received annotations about the presence/absence of chest drains in 1,592 images, created by a board-certified radiologist \cite{oakden2020hidden}. We filter the pneumothorax subset by excluding these expert-annotated images (to use them for evaluation of our annotations), as well as any images of the same patients, leaving a subset of 3,709 chest X-ray images. This selection process is illustrated in Fig.~\ref{fig:flowchart}.

\textbf{\padchest{}} dataset consists of 160,868 chest X-ray images from 67,000 patients, labeled with pathologies but also other findings such as tubes, based on reports associated with the images. Of these reports, 27\% were manually annotated by physicians, and the remaining set was labeled using a supervised Recurrent Neural Network (RNN). We extract a subset of 1,011 images labeled with one of the four tubes shown in Fig.~\ref{fig:chest-xray-tubes}: (a)~chest drain, (b)~tracheostomy, (c)~nasogastric (NSG) and (d)~endotracheal tubes.

\subsection{\textbf{Annotation}}\label{sec:annotation}
Two authors of this paper (CD and TNE) without a medical background provided annotations for 3,709 images from \cxr{} and 1,011 images from \padchest{}, after studying annotator guidelines \cite{macduff2010management,radiology_masterclass,jain2011pictorial}. 

Both annotators first independently labeled the presence or absence of a tube (chest drain for \cxr{}, four types of tubes for \padchest{}). We used the following labels: \{0, 0.25, 0.5, 0.75, 1, INVALID\}, with 0 and 1 indicating absolute certainty about the absence (0) or presence (1). We also included the label `INVALID' due to, for instance, the image not being a frontal chest X-ray. We then combined the raw annotations where both annotators were certain, which was the most prevalent case. This resulted in annotations for 3,543 images for \cxr{} and 1,011 images for \padchest{} (see Fig.~\ref{fig:subsets}).

\subsection{Experimental setup}

We investigate the added value of additional annotations in two main ways. Firstly, we use Cohen's kappa to evaluate the agreement between different annotators. Secondly, we train and evaluate ML models for tube classification tasks using subsets taking into account different annotators or labeling methods, detailed in Section~\ref{sec:results}. We evaluate the Area Under the Receiver Operating Characteristic Curve (AUC) across three model runs, each with a different initialization seed. We report the mean and standard deviation of the AUC over these runs. We always use disjoint subsets of patient-wise images for training and evaluation of the models. 

For the models, we used four different architectures, all with pretrained ImageNet weights: ResNet50~\cite{he2016deep}, InceptionV3~\cite{szegedy2016rethinking},  DenseNet121~\cite{huang2017densely}, and Ours, which is a modified DenseNet121 with three additional layers (pooling, dense and dropout) between the backbone architecture and the flattened layer. 

We performed a grid search to find the best hyperparameters for each setup. For the chest drain classification task with \cxr{}, we use a batch size of 32, a learning rate of 0.0001, and 200 epochs. For the tube classification with \padchest{}, we use a batch size of 32, a learning rate of 0.00001 fine-tuned for 250 epochs. All models use Adam as optimizer and binary cross-entropy as the loss function. Our implementations are based on the Keras library \cite{chollet2015keras}. 

\section{Results} \label{sec:results}
\begin{figure}[t]
    \centering
    \includegraphics[width=0.8\columnwidth]{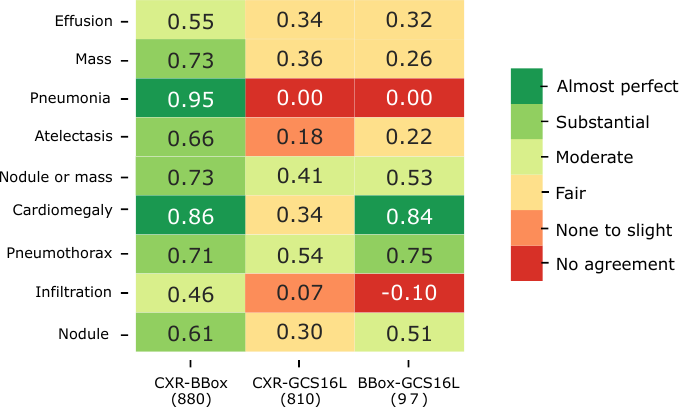}
    \caption{Cohen's Kappa scores of labels from different datasets, all on \cxr{} images. Labels from \cxr{} (`CXR') are from report parsing, the other label sets are from expert image reviews. We include the number of shared images in the parenthesis below the dataset names.}
    \label{fig:KappaPathologies}
\end{figure}

\subsection{Low to moderate agreement on pathology in \cxr{}}

We compare different label sets for pathologies available for \cxr{} using the Cohen's kappa scores in Fig.~\ref{fig:KappaPathologies}. Note that not all combinations of label sets are shown since some did share a few categories while others did not share any images, we report the sample sizes in the same figure.

Overall, we observe almost perfect or substantial agreement for \cxr{} and BBox set, which were released together. Agreements between \cxr{} and three other pathology label sets are at best moderate. Agreement between the other sets varies across pathologies, however, the sample sizes are lower in these cases. The pneumonia class has none to slight agreement, which could be due to the fact that it is a differential diagnosis and requires clinical information beyond X-ray images~\cite{BUSTOS2020101797}. We also find none to slight agreement for emphysema, edema, pleural thickening, consolidation, and fibrosis categories.

\begin{table}[t]
    \centering
    \begin{tabular}{p{1.5cm}K{2cm}K{2cm}K{2cm}}
    Model & \texttt{NonExp} & \texttt{ExpPt1} & \texttt{ExpPt2} \\[0.6ex] 
    \toprule
    \begin{tabular}{@{}l@{}}ResNet \\ Inception \\ DenseNet \\ \textbf{Ours} \end{tabular} & \addstackgap{\alignCenterstack{59.7 \pm& 11.8 \\ 86.1 \pm& 0.3 \\ 90.9 \pm& 0.3 \\ \boldsymbol{92.1} \pm& 1.3}} & 
    \addstackgap{\alignCenterstack{62.4 \pm& 11.5 \\ 85.2\pm& 0.4 \\ 90.8 \pm& 0.1 \\ \boldsymbol{92.6} \pm& 0.2}} & 
    \addstackgap{\alignCenterstack{62.2 \pm& 14.1 \\ 86.2\pm& 0.1 \\ 89.3 \pm& 0.1 \\ \boldsymbol{89.6} \pm& 0.6}}\\ 
    \end{tabular}
    \caption{Chest drain detection (AUC average $\pm$ standard deviation). All models are fine-tuned on \cxr{} training data, and evaluated on disjoint \cxr{} test subsets: our annotations (\texttt{NonExp}), expert annotations part 1 (\texttt{ExpPt1}) and part 2 (\texttt{ExpPt2}). Bold = highest average AUC per column.} 
    \label{tab:chest_drain_cxr}
\end{table}

\subsection{Detector trained with non-expert chest drain annotations generalizes well to expert labels}

We train models for the chest drain detection task using our \texttt{NonExpert} labels, and evaluate on three scenarios:
\begin{itemize}
    \item \textbf{\texttt{NonExp}}: test set of images annotated by two authors of this paper (CD and TNE).
    \item \textbf{\texttt{ExpPt1}} (part 1): 20\% of the expert annotations, provided to the annotators for training purposes at the beginning of this study. 
    \item \textbf{\texttt{ExpPt2}} (part 2): hold-out test set of the expert annotations, provided for evaluation towards the end of this study. 
\end{itemize}

Table~\ref{tab:chest_drain_cxr} shows the mean AUC and standard deviation over three model runs for the backbone models and our proposed detection model (Ours). We observe very similar results across the evaluation scenarios for all the models when considering each model's performance individually. The general good performance on \texttt{ExpPt2} scenario indicates that the models generalize well overall to the hold-out test set with expert annotations.


\subsection{Higher agreement and good generalization with tube annotations in \padchest{}}
\begin{figure}[t]
    \centering
    \includegraphics[width=0.8\columnwidth]{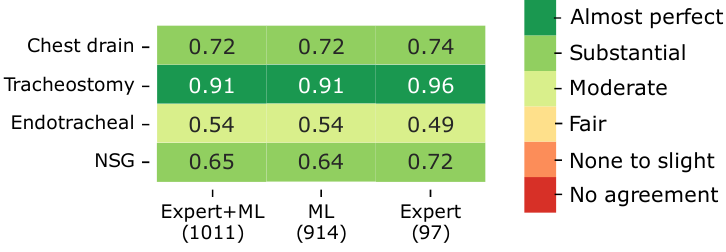}
    \caption{Cohen's Kappa scores between the \texttt{NonExpert} and PadChest annotations (\texttt{Expert+ML}, \texttt{ML}, \texttt{Expert}). We include the number of shared images in the parenthesis below.}
    \label{fig:KappaTubes}
\end{figure}

We consider the annotations of four tube types in \padchest{} dataset: chest drain, tracheostomy, NSG, and endotracheal tube. Fig.~\ref{fig:KappaTubes} shows that our annotations, compared to \padchest{}, present almost perfect agreement ($\kappa=0.91$) on the tracheostomy tube annotations, substantial agreement ($\kappa=0.72$) on the chest drains, and moderate agreement on the remaining two tube types. 

We now use our modified DenseNet121 architecture for the tube classification task. We train the model on the \padchest{} dataset and evaluate it using four different sets of annotations:
\begin{itemize}
    \item \textbf{\texttt{NonExp}}: test set annotated by the authors of this paper (CD and TNE).
    \item \textbf{\texttt{Exp+ML}}: test set in \padchest{} with a tube label.
    \item \textbf{\texttt{ML}}: test subset of images automatically extracted with a RNN (90.4\% of the annotated tube set).
    \item \textbf{\texttt{Exp}}: test subset of images where physicians annotated from the reports (9.6\% of the annotated tube set).
\end{itemize}

Table~\ref{tab:tube_classification_padchest} shows the mean AUC and standard deviation over three model runs for the tube classification task in \padchest{} dataset. Our model evaluated on the \texttt{NonExpert} set shows higher AUC for chest drains and tracheostomy tubes. These two tubes showed a Cohen's kappa that corresponded with almost perfect agreement with \padchest{} original annotations. Moreover, our results suggest that a small number of expert annotations may be more effective than mixing them with automatically extracted annotations. 

\section{Discussion and conclusions} \label{sec:discussion}
\begin{table}[t!]
    \centering
    \begin{tabular}{p{1.5cm}K{2cm}K{2cm}K{2cm}K{2cm}}
    Tube & \texttt{NonExp} & \texttt{Exp+ML} & \texttt{ML} & \texttt{Exp} \\
    \toprule %
    Drain & \textbf{83.7} $\pm$ 4.5 & 77.8 $\pm$ 3.5 &  77.9 $\pm$ 3.5 & 79.0 $\pm$ 4.5  \\
    Trach & \textbf{87.6} $\pm$ 2.0 & 86.4 $\pm$ 2.5 & 83.3 $\pm$ 3.3 & 86.6 $\pm$ 2.5  \\
    Endo. & 74.9 $\pm$ 0.4 & 75.7 $\pm$ 0.1 & 75.7 $\pm$ 0.1 & \textbf{77.1} $\pm$ 2.2  \\
    NSG & 68.4 $\pm$ 1.1 & 74.8 $\pm$ 1.3 & 73.5 $\pm$ 1.7 & \textbf{81.1} $\pm$ 1.8 \\
    \end{tabular}
    \caption{Tube classification AUC $\pm$ standard deviation). We fine-tune on \padchest{} training data and evaluate on \padchest{} test set with our annotations (\texttt{NonExp}), \padchest{} original annotations (\texttt{Exp+ML}) and the two subsets thereof (\texttt{Exp} and \texttt{ML}). Bold = highest average AUC per tube.} 
\label{tab:tube_classification_padchest}
\end{table}

\paragraph{\textbf{Ground truth and task difficulty.}}
Our results show that available pathology labels in a public dataset should not be taken for granted, and not merely seen as benchmarks for ML models. The labels we might consider as ``ground truth'', might not actually reflect the ground truth status of the patient, for example if the labels are automatically (but with errors) extracted from reports, and/or because some diseases require a differential diagnosis. This is in particular reflected in the low agreement for pneumonia, which would typically require additional information next to the chest X-ray \cite{BUSTOS2020101797}.

Given the sometimes low agreement from experts on established pathologies, our results suggest that annotating tubes is an ``easier'' task, since our non-expert annotations show high agreement with those of a certified radiologist, despite the fact that our annotators did not have access to the radiologist's annotations, nor any contact with the radiologist. While this is a small-scale study, we believe non-expert annotations for such ``easier'' tasks could still provide additional value in exploring different strategies for training pathology detection models.     

\paragraph{\textbf{Training strategies for non-expert annotations.}}
There are several strategies how non-expert annotations could be used to improve ML training, here we describe three possibilities. 
The first strategy is to incorporate the annotations into a multi-task learning setup. The additional annotations act as a regularizer, since the model is trained to both predict both the diagnosis and the presence of a visual element, like chest tubes. Such regularization helps the model learn better pathology representations. A study on the classification of skin lesions, which included non-expert annotations of features such as ``asymmetry'' (a more intuitive feature than a diagnosis), showed that using these non-expert annotations with multi-task learning improved the model performance \cite{raumanns2021enhance}. We have conducted similar, preliminary, experiments with chest X-ray classification with our additional annotations. Our early results indeed tend to show increased performance, but would still require further validation, so we have chosen not to focus on this aspect in this paper.

A second strategy for using additional annotations is to avoid shortcut learning in an adversarial \cite{zhang2018mitigating} setup, where the model optimizes the learning towards features that are discriminative for the pathology while minimizing the effect of the shortcut. This strategy is similar to work on bias and fairness, where the model learns to minimize the effect of a protected attribute, like sex or age, on the model, for example \cite{abbasi2020risk,adeli2021representation}.

A third strategy could be based on contrastive learning \cite{khosla2020supervised}, where pairs of positive (similar) and negative (dissimilar) samples can guide the model towards capturing the discriminative predictive features. Additional annotations could be used here to define or refine the (dis)similarity functions.

\paragraph{\textbf{Limitations.}}
A limitation of our study is that the authors who labeled the images (with a data science but no medical background) dedicated significant effort to understanding the annotation task. This included reading medical literature and tutorials \cite{macduff2010management,radiology_masterclass,jain2011pictorial}. As a result, our findings might not be applicable to other non-expert annotators, such as those on crowdsourcing platforms, where due to the financial incentives annotators would likely take on a variety of annotation tasks (not just medical images), and spend less time on each specific task.  

\paragraph{\textbf{Concluding remarks.}}
Our work is in spirit close to other studies that add additional information to existing datasets, for example in the form of additional documentation (e.g., Datasheets \cite{gebru2021datasheets} or the later introduced Healthsheet \cite{rostamzadeh2022healthsheet}), additional annotations, segmentation masks \cite{gaggion2024chexmask}, or experimental results that show evidence of shortcuts \cite{oakden2020hidden,JimenezSanchez2023Shortcuts,gichoya2022ai,degrave2021ai,winkler2019association}. These extensions of existing datasets have been referred to as ``research artifacts'' \cite{jimenez2025picture}. As a community aiming to build upon previous work, we recommend that we not simply treat datasets ``as-is'', but thoughtfully consider such existing research artifacts when interpreting and drawing conclusions from our results.

To conclude, while our work shows that non-expert annotations could be advantageous in the training stage, we want to emphasize that there is \emph{no free lunch} for evaluation, and one should never use such annotations to claim superiority of one algorithm over another. For reliable evaluation of models, we always need reliable ground truth, more diverse datasets, and expert involvement. 

%
%
\bibliographystyle{splncs04}
\bibliography{refs,refs_veronika}

\clearpage
\appendix
\counterwithin{figure}{section}
\counterwithin{table}{section}
\renewcommand\thefigure{\thesection\arabic{figure}}
\renewcommand\thetable{\thesection\arabic{table}}

\section{Supplementary Material}

\begin{figure}[h]
    \centering
    \includegraphics[width=0.55\columnwidth]{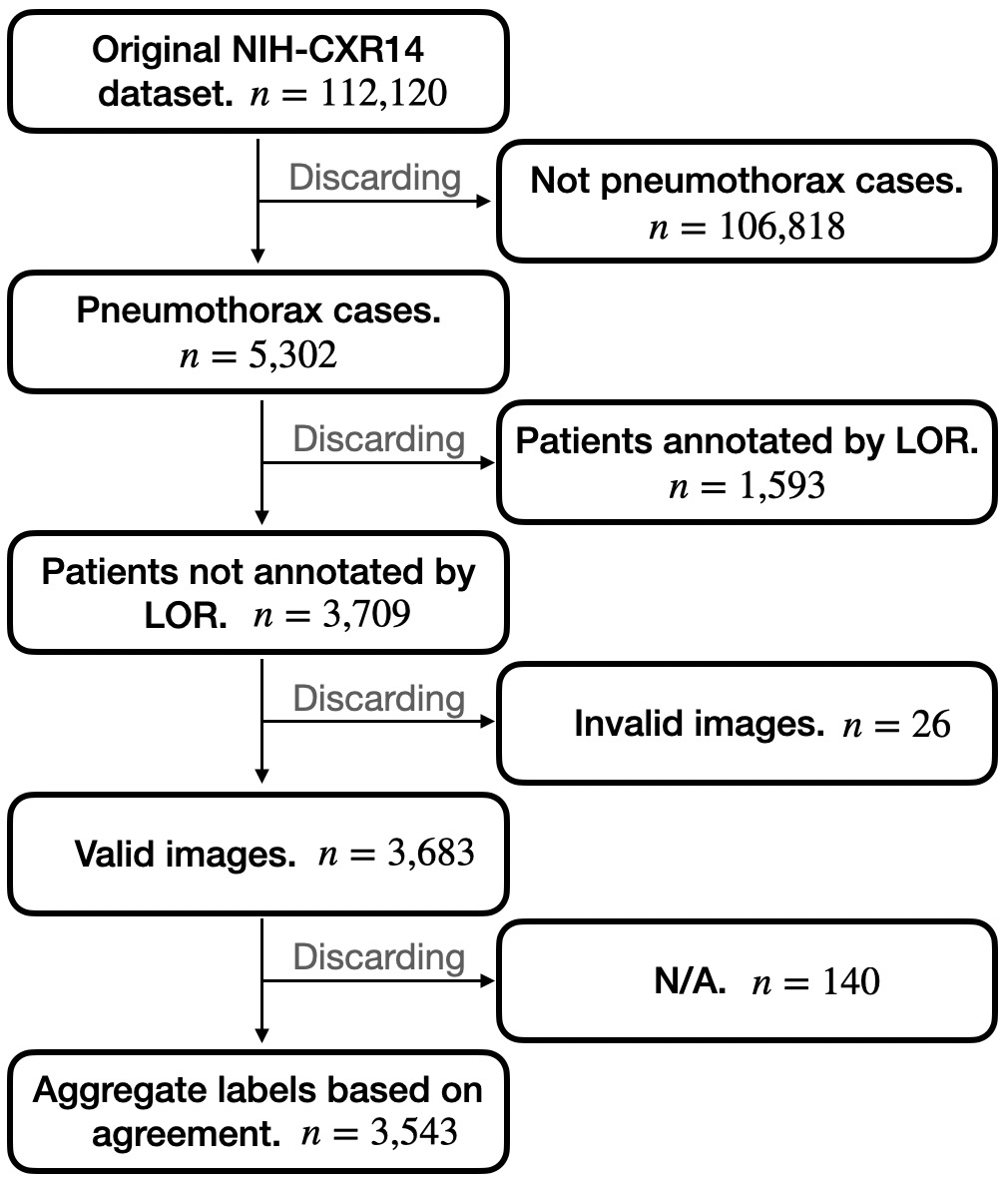}
    \caption{Flowchart of \cxr{} data extraction and annotation process. LOR refers to the radiology expert, $n$ is the number of images at that step. }
    \label{fig:flowchart}
\end{figure}

\renewcommand{\arraystretch}{1.5}

\begin{table}[ht!]
    \centering
    \resizebox{\textwidth}{!}{  
    \begin{tabular}{p{2.5cm}K{2.2cm}K{2.5cm}K{2.2cm}p{6cm}}
    \toprule
     \textbf{Dataset} & \textbf{Dataset size} & \textbf{Labeling method} & \textbf{No. labels} & \textbf{Label note} \\ [0.5ex]
    \midrule
    \padchest{} \cite{BUSTOS2020101797} & 160,868 & Report review + Report parsing & 193 &  \\
    \midrule
    \cxr{} \cite{chest-xray8} & 112,120 & Report parsing & 15 & Includes a `No Finding' label \\
    BBox \cite{chest-xray8} & 880 & Image review & 8 & Contains the original 8 labels from \cxr{} \\
    GCS16L \cite{CHX14_alllabels} & 810 & Image review & 16 & Includes 14 labels from \cxr{} and `Abnormal' and `Other' \\
    GCS4L \cite{CHX14_fourlabel} & 4,376 & Image review & 4 & Labels: `Airspace opacity', `Pneumothorax', `Module/mass' and `Fracture' \\ 
    RSNA \cite{RSNA} & 26,684 & Image review & 2 & Labels: `No pneumonia' and `Pneumonia' \\
    \bottomrule
    \end{tabular}
    }
    \caption{Chest X-ray datasets, their sizes, labeling methods, and information about their labeled pathologies. BBox, GCS16L, GCS4L and RSNA are additional annotation sets for \cxr{}.}
    \label{tab:datasets-labels}
\end{table}

\end{document}